\documentclass[conference]{IEEEtran}
\usepackage{microtype}
\usepackage{booktabs}
\usepackage{graphicx} 
\usepackage[utf8]{inputenc}
\usepackage{tikz}
\usepackage{xcolor} 
\usetikzlibrary{shapes.geometric, arrows.meta, positioning}
\usepackage{lscape}
\usepackage{multirow}
\usepackage{colortbl}
\usepackage{cite}
\usepackage{amsmath,amssymb,amsfonts}
\usepackage{algorithmic}
\usepackage{textcomp}
\usepackage{xurl}
\usepackage{hyperref}

\def\BibTeX{{\rm B\kern-.05em{\sc i\kern-.025em b}\kern-.08em
    T\kern-.1667em\lower.7ex\hbox{E}\kern-.125emX}}

\newcommand\copyrighttext{%
  \footnotesize \textcopyright~2026 IEEE. Personal use of this material is permitted.
  Permission from IEEE must be obtained for all other uses, in any current or future
  media, including reprinting/republishing this material for advertising or promotional
  purposes, creating new collective works, for resale or redistribution to servers or
  lists, or reuse of any copyrighted component of this work in other works.}
\newcommand\copyrightnotice{%
\begin{tikzpicture}[remember picture,overlay]
\node[anchor=south,yshift=8pt] at (current page.south) {\parbox{\dimexpr\textwidth\relax}{\copyrighttext}};
\end{tikzpicture}%
}

\makeatletter
\newcommand{\linebreakand}{%
  \end{@IEEEauthorhalign}
  \hfill\mbox{}\par
  \mbox{}\hfill\begin{@IEEEauthorhalign}
}
\makeatother

\begin{document}

\title{Lines and Ladders: A Context-Aware Multi-Agent Framework for Large-Scale Retail Price Taxonomy}
\author{
\IEEEauthorblockN{Ravi Teja Chunduri}
\IEEEauthorblockA{\textit{Walmart Global Tech}\\
Bentonville, AR, USA \\
raviteja.chunduri@walmart.com}
\and
\IEEEauthorblockN{Srikaran Reddy Boya}
\IEEEauthorblockA{\textit{Walmart Global Tech}\\
Bentonville, AR, USA \\
Srikaran.Reddy.Boya@samsclub.com}
\and
\IEEEauthorblockN{Deep Narayan Mishra}
\IEEEauthorblockA{\textit{Walmart Global Tech}\\
Sunnyvale, CA, USA \\
deep.mishra@walmart.com}

\linebreakand 

\IEEEauthorblockN{Ajay Kumar B}
\IEEEauthorblockA{\textit{Walmart Global Tech}\\
Seattle, WA, USA \\
Ajay.Kumar10@walmart.com}
\and
\IEEEauthorblockN{Karthik Kumaran}
\IEEEauthorblockA{\textit{Walmart Global Tech}\\
Sunnyvale, CA, USA \\
karthik.kumaran@walmart.com}
\and
\IEEEauthorblockN{Pranay Kona}
\IEEEauthorblockA{\textit{Walmart Global Tech}\\
Sunnyvale, CA, USA \\
pranay.kona@walmart.com}
}

\maketitle
\copyrightnotice
\begin{abstract}
Maintaining price consistency and executing an Every Day Low Price strategy is critical for global retailers. However, with catalogs spanning millions of active items, manual governance of price relationships is infeasible. Inconsistent pricing across item variants distorts customer value perception and cannibalizes sales. To address this, we present a scalable, context-aware Multi-Agent Framework designed to automate the construction of ``Lines and Ladders'' pricing taxonomies. Our framework employs specialized LLM agents to construct these coherent pricing structures by identifying key attributes, extracting multi-modal values, and applying hierarchical grouping logic. Evaluated on real-world enterprise data and deployed in production, our 3-Agent system achieves an F1-score of 0.83 for Lines, outperforming single-agent baselines by mitigating cognitive overload. The system achieves $>90\%$ precision and $>75\%$ recall in Food \& Consumables, and $80.2\%$ assignment accuracy in the unstructured General Merchandise catalog.
\end{abstract}


\section{Introduction}
Retail pricing directly dictates the core profitability of a retailer \cite{profitability_nagle}. To maintain consistency, retailers structure products into ``lines and ladders''—a logical hierarchy grouping items by features, quality, and business context \cite{monroe2003pricing}. Traditionally, these structures are created through a manual process led by experts, where category managers make decisions based on deep domain knowledge \cite{basroy2001impact}. However, this approach is inherently static and does not meet the dynamic nature of modern
retail catalogs.

For global retailers, price governance complexity scales non-
linearly. With millions of active items, high assortment velocity
renders manual oversight infeasible. The core problem is not merely the volume of items, but the operational bottleneck created by the inconsistent and often non-standardized nature of catalog data. When relationships are ignored, illogical pricing gaps emerge, leading to customer arbitrage \cite{xia2004price} \cite{krishna_2002}. In un-managed catalogs, we frequently observe two gaps: \textbf{Variant Inconsistency} (e.g., identical bikes priced at \$238 vs \$199 solely due to color) and \textbf{Data Discrepancy} (e.g., a 0.1-inch typo in a mirror's dimensions creating a \$7.50 gap). 

To address this scalability challenge, we propose a context-aware multi-agent framework. Unlike traditional rule-based systems, our approach leverages specialized agents to automatically cluster products into meaningful ``Lines \& Ladders''.
This system automates the identification and extraction of key attributes, customized for each ProductType. We utilize this architecture to simulate and replicate the decision-making logic of a human expert at scale \cite{park2023generative}. 

This framework serves the primary objective of automated price governance, establishing the structural foundation required for next-generation autonomous pricing systems such as algorithmic demand forecasting and dynamic pricing models \cite{fader1996modeling}. The main contributions of this paper are summarized as follows:
\begin{itemize}
\item \textbf{Formalizing Taxonomy Construction:} We decompose pricing governance into Similarity \& Variance discovery via a 3-agent architecture.
\item \textbf{Scalable Multi-Modal Execution:} We detail a distributed pipeline utilizing a multi-model routing strategy to process millions of items efficiently.
\item \textbf{Validating Industrial Efficacy:} We provide extensive ablation studies and post-launch metrics demonstrating the system's scalability and operational impact in a live enterprise environment.
\end{itemize}

\section{Related Work}
The concept of ``lines and ladders'' has been indirectly established in retail strategy for decades. Dean \cite{dean1950problems} originally defined ``price lining'' as the practice of offering a range of products at specific, distinct price points to signal varying levels of quality to the consumer. While there is extensive literature on product line pricing problems, as comprehensively reviewed by Guiltinan \cite{guiltinan2011progress}, our work addresses a fundamentally different challenge. Unlike traditional approaches that focus on the mathematical optimization of setting prices, our objective is to automatically cluster a massive catalog into coherent product lines based on quality and features. This structural organization is a prerequisite for scalable price governance rather than price setting itself.

The term ``ladder'' is a more recent term and can be interpreted in various ways. Dobbs \cite{dobbs2016does} describes a ladder as a form of wholesale pricing contract where the wholesale price is contingent on the retail price, facilitating tiered price setting by the manufacturer. In a marketing context, price laddering aligns closely with price lining, typically involving a ``good-better-best'' strategy \cite{mohammed2005art}. Furthermore, the concept of ``climbing a ladder'' refers to the strategy of incentivizing customers to upgrade to higher-quality products within the offered set, as described by \cite{tgndata_nd_ladder} and \cite{pricebeam2017ladder}.

Draganska and Jain \cite{draganska2006consumer} further distinguish between vertical and horizontal product differentiation. In vertical differentiation, products vary by quality, whereas in horizontal differentiation, firms offer products that vary in characteristics such as scent, color, or flavor rather than quality. Their empirical analysis confirms that product lines function as effective price discrimination tools, justifying the common strategy of pricing lines differently based on quality while pricing variants (e.g., flavors or colors) uniformly. While the literature offers various definitions for ladders, in the context of large-scale retail, we specifically define ladders in terms of volume incentives. Therefore, for the purposes of this paper, we refer to ``volume discount ladders'' simply as ladders.

Although these concepts explain the strategic ``why'', they fail to address the operational ``how'' for massive, unstructured datasets. Pre-LLM (Large Language Models) approaches relied on structured tabular data and extensive feature engineering \cite{mudgal2018deep}, a workflow rendered impractical by the heterogeneity of modern retail catalogs. The manual effort required to standardize attributes across millions of items is practically impossible. LLMs offer a paradigm shift, enabling the direct interpretation of unstructured text and images to automate feature extraction at scale.

\section{Methodology}
To overcome the limitations of manual governance, we propose a Context-Aware Multi-Agent Framework. We avoid a single-prompt approach as it yields unstable outputs. Instead, the system leverages specialized LLMs to perform a two-pronged analysis, discovering key attributes from non-standardized data.

\subsection{Problem Formulation \& Scoping}
\begin{figure}[h]
\centering
\includegraphics[width=3.35in, trim=0pt 0pt 0pt 0pt, clip]{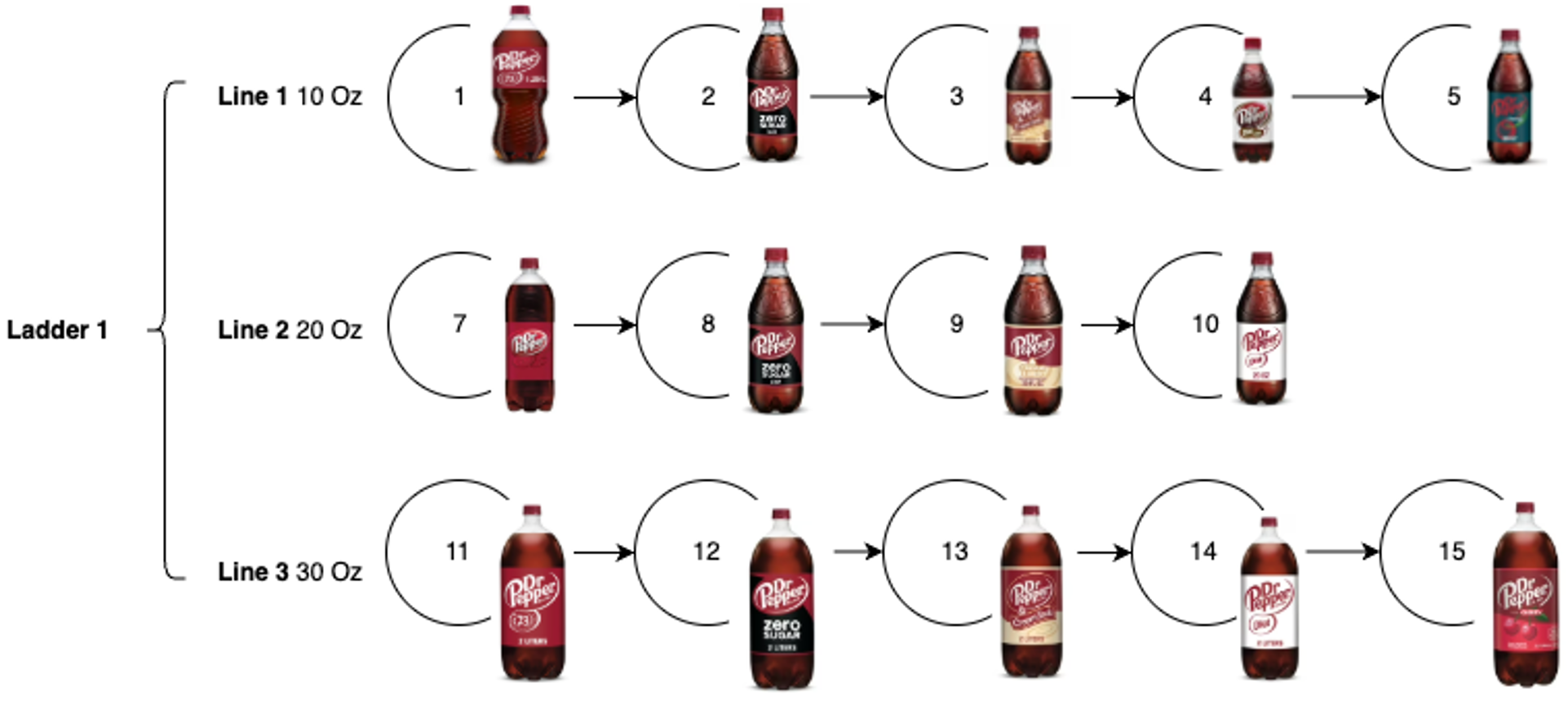}
\caption{Definition of Lines \& Ladders}
\label{fig:definition}
\end{figure}

Let $C = \{i_1, i_2, ..., i_N\}$ be a catalog of $N$ items, each with raw metadata $R_i$ (e.g., title, description, image). As shown in Figure \ref{fig:definition}, we map each item to a \textbf{Line ($L_k$)}: a subset of items sharing functional equivalence (e.g., identical items differing only in flavor or color), where $\forall i_a, i_b \in L_k \implies Price(i_a) = Price(i_b)$ and Lines aggregate into a \textbf{Ladder ($M_j$)}: a collection of Lines linked by value-based differentiation (e.g., volume, pack size), enforcing a tiered pricing logic where $Volume(L_a) > Volume(L_b) \implies UnitPrice(L_a) < UnitPrice(L_b)$.

Analyzing millions of items via a global, top-down approach is computationally expensive and lacks context. Therefore, we decompose the catalog using \textbf{ProductType} (e.g., ``Candles'') as our granular anchor. Scoping the analysis to items with shared structural characteristics ensures the extracted attributes are highly relevant, allowing the framework to generate precise, tailored taxonomies rather than relying on generic, one-size-fits-all rules.

\subsection{Dual-Pronged Attribute Discovery}
Constructing effective taxonomies requires distinguishing between \textbf{Similarity Attributes} (features defining functional equivalence at the same price point) and \textbf{Variance Attributes} (features driving price differentiation). We achieve this using two specialized LLM agents. To ensure the generated taxonomy is strictly based on physical attributes, these agents operate without few-shot examples of historical taxonomies. Because legacy merchant structures often reflect ad-hoc operational strategies (e.g., grouping all items from a single supplier) rather than strict attribute equivalence, injecting them into the prompts would introduce strategic noise and confuse the algorithmic extraction.

\subsubsection{Identifying Price Similarity Attributes (Agent 1)}
This agent isolates base features common to identically priced items, defining the core value proposition. Figure \ref{fig:agent1_strategy} outlines the prompt strategy. 
\begin{itemize}
\item \textbf{Density-based Sampling:} From \texttt{ProductType} data set $D_{pt}$, we identify the $K$ most frequent tuples $T_{sim} = \{(Brand, MSRP)_{1}, ..., (Brand, MSRP)_{K}\}$ where $MSRP$ represents manufacturer’s suggested retail price. We strictly utilize base MSRP to avoid promotional noise. 
\item \textbf{Corpus Generation:} For each tuple, we sample a corpus $C_t$ of 15 item descriptions, yielding $K$ distinct corpora representing clusters of identically priced items. 
\item \textbf{Iterative Extraction:} Agent 1 processes each $C_t$ to extract consistent core attributes, utilizing a persistent memory $P_{sim}$ to iteratively refine its logic. For a ``Candles'' \texttt{ProductType}, outputs might include \textsf{\{weight\_oz, height\_inches, burn\_time\_hours\}}.
\end{itemize}

\begin{figure}[t]
\centering
\small
\begin{tabular}{|p{0.95\linewidth}|}
\hline
\textbf{Objective:} Extract quantifiable attributes justifying identical pricing. \\
\textbf{Constraints:} \\
1. \textbf{Quantifiable Only:} Reject categorical features; focus on measurable specs (e.g., weight, capacity). \\
2. \textbf{Consistency:} Attribute values must be invariant across the group. \\
3. \textbf{Evidence Tiering:} \\
\quad $\bullet$ \textit{Tier 1 (Explicit):} Stated in text $\rightarrow$ \textbf{Accept}. \\
\quad $\bullet$ \textit{Tier 2 (Inferred):} Logically derived $\rightarrow$ \textbf{Accept}. \\
\quad $\bullet$ \textit{Tier 3 (Speculative):} No evidence $\rightarrow$ \textbf{Reject}. \\
4. \textbf{Standardization:} Enforce output schema and reuse names from persistent memory. \\
\hline
\end{tabular}
\caption{\textbf{Prompt strategy to identify similarity attributes}}
\label{fig:agent1_strategy}
\end{figure}

\subsubsection{Identifying Price Variance Attributes (Agent 2)}
This agent identifies premium features that justify price stratification. Figure \ref{fig:agent2_strategy} outlines the prompt strategy, instructing the agent to act as a Pricing Strategist.
\begin{itemize}
\item \textbf{Stratified Sampling:} To capture price diversity, we select $B_{top}$, the top 5 frequent brands in $D_{pt}$. For each brand $b \in B_{top}$, we identify its top 5 frequent MSRP points, creating a stratified set $T_{var}$ of up to 25 groups, $T_{var} = \{(Brand_{1}, MSRP_{1}),$ $\dots, (Brand_{5}, MSRP_{5})\}$. Empirical trials indicated that expanding beyond 5 brands introduces excessive nomenclature variance that obscures core \texttt{ProductType} signal.
\item \textbf{Corpus Generation:} We sample 5 items for each group in $T_{var}$. Unlike Agent 1, Agent 2 ingests the entire stratified dataset $D_{var}$ as a single input to analyze a wide quality and price spectrum.
\item \textbf{Comparative Analysis:} Agent 2 compares low- and high-priced groups across $D_{var}$ to identify differentiating features (e.g., \textsf{\{number\_of\_wicks, wax\_type, has\_decorative\_vessel\}}). The agent is also equipped with a memory module, allowing it to incorporate feedback and refine attributes based on specific business requirements.
\end{itemize}

\begin{figure}[b]
\centering
\small
\begin{tabular}{|p{0.95\linewidth}|}
\hline
\textbf{Objective:} Compare price-stratified groups to identify differentiating attributes. \\
\textbf{Analysis Framework:} \\
1. \textbf{Inventory \& Evidence:} List attributes with evidence tiers as shown in Figure \ref{fig:agent1_strategy}; Reject speculative (Tier 3) evidence. \\
2. \textbf{Type Classification:} \\
\quad $\bullet$ \textit{Type 1 (Differentiating):} Consistent within group, varies across price tiers (Primary). \\
\quad $\bullet$ \textit{Type 2 (Foundational):} Essential specs regardless of price impact. \\
3. \textbf{Selection:} Finalize attributes that explain price deltas or define identity. \\
\textbf{Constraints:} Standardized naming; no generic terms (e.g., ``style''); Follow output schema. \\
\hline
\end{tabular}
\caption{\textbf{Prompt strategy to identify variance attributes}}
\label{fig:agent2_strategy}
\end{figure}

\begin{figure*}[!t]
\centering
\includegraphics[width=6.50 in, trim=0pt 7pt 0pt 8pt, clip]{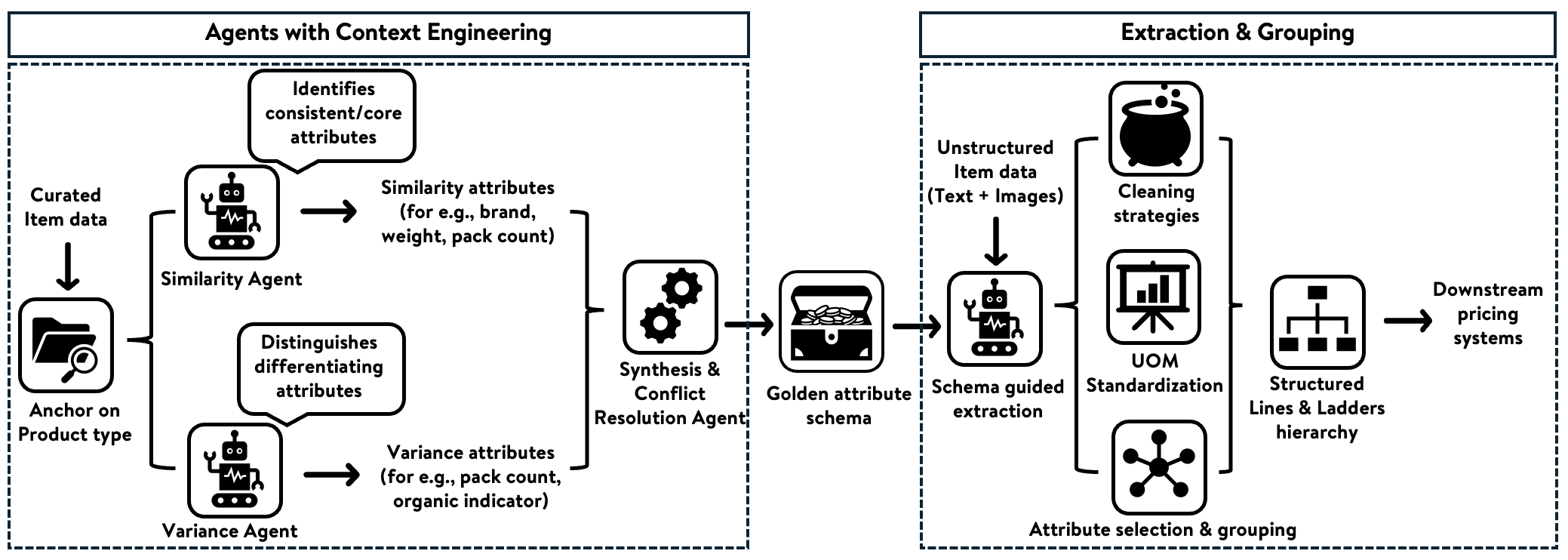}
\caption{Implemented methodology for each \texttt{ProductType}}
\label{fig:pipeline_complete}
\end{figure*}

Figure \ref{fig:pipeline_complete} illustrates this end-to-end workflow, detailing the parallel execution and subsequent synthesis of Agents outputs.

\subsubsection{Parameter Selection \& Heuristics} 
The sampling constants used by Agent 1 (30 groups / 15 items) and Agent 2 (25 groups / 5 items) were derived via parameter tuning to optimize the Pareto frontier between signal-to-noise ratio and LLM token costs. As shown in Table \ref{tab:heuristics_ablation}, providing more data to the LLM degrades performance. Expanding sample breadth (e.g., 80 groups) introduces long-tail catalog noise and niche brands that confuse schema generation. Expanding depth (e.g., 30 items) causes context bloat, leading the LLM to hallucinate variance attributes based on minor, irrelevant differences. Furthermore, average pipeline runtimes were scaled up to 155 minutes for the 80/30 configuration, compared to $\sim$65 minutes for our proposed thresholds.



\begin{table}[t] 
\centering
\caption{Impact of Sample Size (Groups / Items per Group)} 
\label{tab:heuristics_ablation}
\resizebox{\columnwidth}{!}{%
\begin{tabular}{lccccc}
\toprule
\textbf{Metric (Avg 14 Categories)} & \textbf{50/15} & \textbf{80/15} & \textbf{50/30} & \textbf{80/30} & \textbf{Proposed$^*$} \\
\midrule
\textbf{Lines (F1 Score)} & 0.77 & 0.75 & 0.78 & 0.77 & \textbf{0.83} \\
\textbf{Ladders (F1 Score)} & 0.80 & 0.76 & 0.78 & 0.75 & \textbf{0.82} \\
\bottomrule
\multicolumn{6}{l}{\rule{0pt}{3ex}\footnotesize $^*$30 groups / 15 items for Agent 1, and 25 groups / 5 items for Agent 2.}
\end{tabular}%
}
\end{table}

\subsection{Synthesis Agent}
The similarity and variance agents yield distinct attributes sets, $A_{sim}$ and $A_{var}$, which often exhibit semantic overlap. To resolve these dualities, a Synthesis Agent performs a mapping function $\Phi$ to produce a canonical schema $S_{pt}$: $ S_{pt} = \Phi(A_{sim} \cup A_{var}) $

Figure \ref{fig:agent3_strategy} provides the implemented prompt logic. The execution of $\Phi$ ensures schema integrity through three sequential operations. 
\begin{itemize}
\item \textbf{Attribute Consolidation \& Semantic Resolution:} Raw outputs from Agents 1 and 2 are aggregated to eliminate redundancy, resolve synonymy (e.g., merging ``flavor'' and ``taste''), and enforce standardized naming conventions. 
\item \textbf{Attribute-Unit Decoupling:} To resolve UOM inconsistencies (e.g., oz vs g) and enable comparative analysis, the framework explicitly decouples numerical values from units, defining a scalar value field and a UOM field for every quantitative attribute. This yields a streamlined, high-fidelity schema blueprint for downstream extraction.
\end{itemize}

\begin{figure}[ht]
\centering
\small
\begin{tabular}{|p{0.95\linewidth}|}
\hline
\textbf{Objective:} Synthesize Similarity and Variance lists into a canonical schema. \\
\textbf{Analysis Framework:} \\
1. \textbf{Consolidate:} Merge lists; resolve synonyms (e.g., ``watts'' $\rightarrow$ ``motor\_power'') using similarity output as precedence. \\
2. \textbf{Filter:} Remove global attributes (Brand, Color) and generic terms (e.g., ``features''). \\
3. \textbf{UOM Decoupling:} For every physical metric, generate a companion \textsf{\_uom} field. \\
4. \textbf{Schema Definition:} Define data types, defaults, and extraction prompts. \\
\textbf{Constraints:} Standardized naming; Follow output schema; Justify high-cardinality attributes.\\
\hline
\end{tabular}
\caption{\textbf{Prompt strategy for Schema Synthesis}}
\label{fig:agent3_strategy}
\end{figure}

\subsection{Multi-Modal Attribute Extraction}
Following schema synthesis, the framework populates the schema for every item to transform unstructured data into structured feature vectors. 

\subsubsection{Schema Definition:} The canonical schema $S_{pt}$ generated in the previous step consists of $k$ attributes $\{a_1, a_2, ..., a_k\}$. Each attribute $a_j$ is formally defined as a tuple $a_j = \langle \textit{name},$ $\textit{data type},$ $\textit{default},$ $\textit{prompt} \rangle$.
Figure \ref{fig:Sample_Schema} shows examples of the generated attribute schema for the ``Candles'' \texttt{ProductType}.

\begin{figure}[t]
\centering
\small
\begin{tabular}{|p{0.9\linewidth}|}
\hline
$\bullet$ \textbf{Attribute Name:} wax\_type \\
\textbf{Attribute Type:} String \\
\textbf{Default value:} paraffin \\
\textbf{Prompt:} Identify the type of wax used in the candle (e.g., soy, paraffin, beeswax, coconut) from the image or description. \\

$\bullet$ \textbf{Attribute Name:} is\_decorative \\
\textbf{Attribute Type:} Boolean \\
\textbf{Default value:} False \\
\textbf{Prompt:} Indicate True if the candle features specific artwork, themes, or a sculpted shape. \\
\hline
\end{tabular}
\caption{\textbf{Sample Schema Definition for Extraction}}
\label{fig:Sample_Schema} 
\end{figure}
\subsubsection{Multi-Modal Execution:} We employ a multi-modal function $f_{LLM}$ that maps an item's raw data $R_i$ (text + images) to a feature vector $V_i$ based on the schema $S_{pt}$: $ V_i = f_{LLM}(R_i, S_{pt}) $

\begin{figure}[b]
 \centering
\includegraphics[width=0.45\textwidth, trim=0pt 7pt 0pt 5pt, clip]{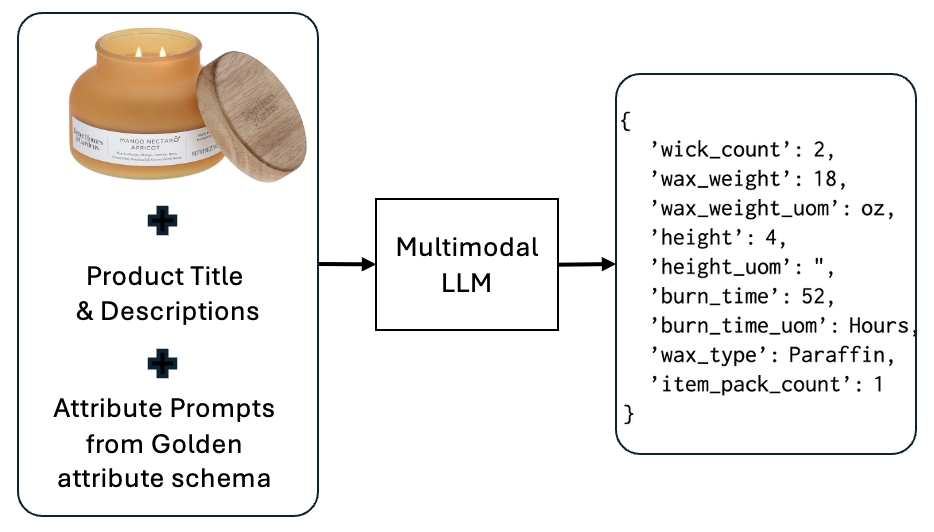}
 \caption{Implemented methodology for attribute extraction}
 \label{fig:attr_extraction} 
\end{figure}

This multi-modal capability is critical for retail catalogs, where key specifications (e.g., ``4-pack'' or ``organic'') often appear only on packaging images. Figure \ref{fig:attr_extraction} depicts this workflow, where the model ingests item context and schema prompts to generate structured values. Unlike traditional methods requiring thousands of specific NER models \cite{chen-etal-2023-named}, this approach utilizes a single generalized agent to extract arbitrary attributes defined by the dynamic schema, resolving the scalability bottleneck.

Because traditional pre-sanitization (e.g., regex) fails on missing or unstructured catalog data, this Multi-Modal Extractor acts as the sanitization engine itself. Initial experiments with smaller open-weight models (e.g., Nemotron VLM 2B/8B/12B) failed to achieve production accuracy without large-scale labeled data tailored to our complex schema, necessitating our reliance on frontier models.

\subsection{Data Standardization and Normalization}
Raw extraction yields structured data, but inconsistencies impede grouping. We employ a post-processing pipeline $\Psi$ to normalize the extracted feature vector $V_i$.
\begin{itemize}
\item \textbf{Numerical Standardization (UOM):} For attributes suffixed with \textsf{\_uom} (e.g., weight), a conversion function $\phi_{uom}$ transforms the raw value $v_{raw}$ and unit $u_{raw}$ into a normalized value $v_{norm}$ in a standard base unit $u_{base}$:
$$ v_{norm} = \phi_{uom}(v_{raw}, u_{raw}) \rightarrow u_{base} $$
For example, $(12, \text{lbs}) \rightarrow 192$ (oz). We enforce a strict $5\%$ numerical tolerance to absorb floating-point inaccuracies during UOM conversions, preventing the artificial separation of identical items.
\item \textbf{Categorical Semantic Clustering:} To resolve semantic fragmentation (e.g., ``Soy'' vs. ``Soy Blend''), an LLM-based clustering function $C_{sem}$ maps a raw value $u_j$ from the set of unique values $U$ to a canonical term $u_{canon}$, preventing artificial fragmentation: $ u_{canon} = C_{sem}(u_j | U) $
\item \textbf{Brand Normalization:} To strictly govern brand identity, a hybrid function $R_{brand}$ reconciles the extracted brand $B_{ext}$ with the catalog master $B_{int}$ via sub-string mapping and a safety rail based on Jaccard Similarity ($J$) and brand frequency ($b_f$):
$$ B_{final} = \begin{cases} B_{int} & \text{if } J(B_{ext}, B_{int}) = 0 \land b_f(B_{int}) \ge 5 \\ B_{ext} & \text{otherwise} \end{cases} $$
Here, $J(B_{ext}, B_{int})$ denotes the Jaccard Similarity between the extracted and internal brand strings, and $b_f(B_{int})$ represents the frequency count of the internal brand within the catalog. This prioritizes the system of record $B_{int}$ when the extracted brand is disjoint from a well-established entity, preventing hallucinated overrides.
\end{itemize}

\subsection{Adaptive Taxonomy Construction}
Grouping is scoped at the brand level to respect attribute applicability, as brands within a \texttt{ProductType} often employ distinct conventions (e.g., weight vs. burn time).

\subsubsection{Feature Selection \& Pruning} 
We apply heuristic filters on the normalized vector $V_i$ to retain discriminative features:
\begin{itemize}

  \item \textbf{Sparsity \& Imputation:} Through iterative deployment and experimentation across diverse catalog samples, we established operational heuristics to drop attributes with $>65\%$ nullity or $>70\%$ cardinality; retaining sparser features consistently caused artificial fragmentation of product lines. Gaps in retained attributes are imputed using $S_{pt}$ defaults.

  \item \textbf{Numerical Binning:} To handle numerical variance, we apply $\epsilon$ neighborhood clustering. Dictated by strict business requirements, values within a $5\%$ tolerance (e.g., 12.0 oz vs 12.3 oz) are binned into a single discrete value $v_{bin}$.
\end{itemize}

\subsubsection{Hierarchical Grouping Logic}
Let $A_{cat}$ be the set of categorical attributes and $A_{num}$ be the set of numerical/volume attributes.
\begin{itemize}
  \item \textbf{Line Generation:} Items are grouped by the complete set $A_{total} = A_{cat} \cup A_{num}$. Each unique combination forms a Line ($L_k$) assigned a persistent UUID:
  $$ L_k = \{i \in C \mid V_i(A_{total}) = \text{const}\} $$
  \item \textbf{Ladder Generation:} Relaxing constraints by excluding $A_{num}$, we group by $A_{cat}$ to cluster Lines into Ladders ($M_j$), linking items differing only by size or pack quantity:$$ M_j = \{i \in C \mid V_i(A_{cat}) = \text{const}\} $$
\end{itemize}
Consequently, a ladder is defined as the union of its constituent Lines: $M_j = \bigcup_{k} L_k$. This logic automatically structures the catalog, placing functionally equivalent items into Lines and connecting them via volume-based Ladders.

\subsection{Human-in-the-Loop Feedback Mechanism}
While the multi-agent framework provides a robust baseline, the inherent stochasticity of LLMs and the specificity of business strategy (e.g., pricing soft drinks at the parent brand level) necessitate human intervention. We incorporate a Human-in-the-Loop (HITL) mechanism to capture merchant corrections as high-quality labeled signals. This feedback updates a \texttt{ProductType} specific contextual memory, $P_{context}$, designed to drive future pipeline refinements. 
\begin{itemize}
\item \textbf{Strategic Alignment:} Agents 1 and 2 will learn to prioritize or ignore attributes based on merchant preferences. 
\item \textbf{Schema Refinement:} Agent 3 will update its merging logic to respect business-specific naming conventions and optimize extraction prompts. 
\end{itemize}

As $P_{context}$ accumulates, agents recall these corrections to ensure subsequent runs align with established business logic, allowing the system to progressively evolve into a domain-adapted expert. While this HITL architecture successfully captures immediate strategic intent, quantifying the system's long-term convergence toward expert-level behavior remains an ongoing area of empirical validation within our deployed environment.

\section{Deployment at Scale}
\begin{figure*}[t]
\centering
\includegraphics[width=6.5in]{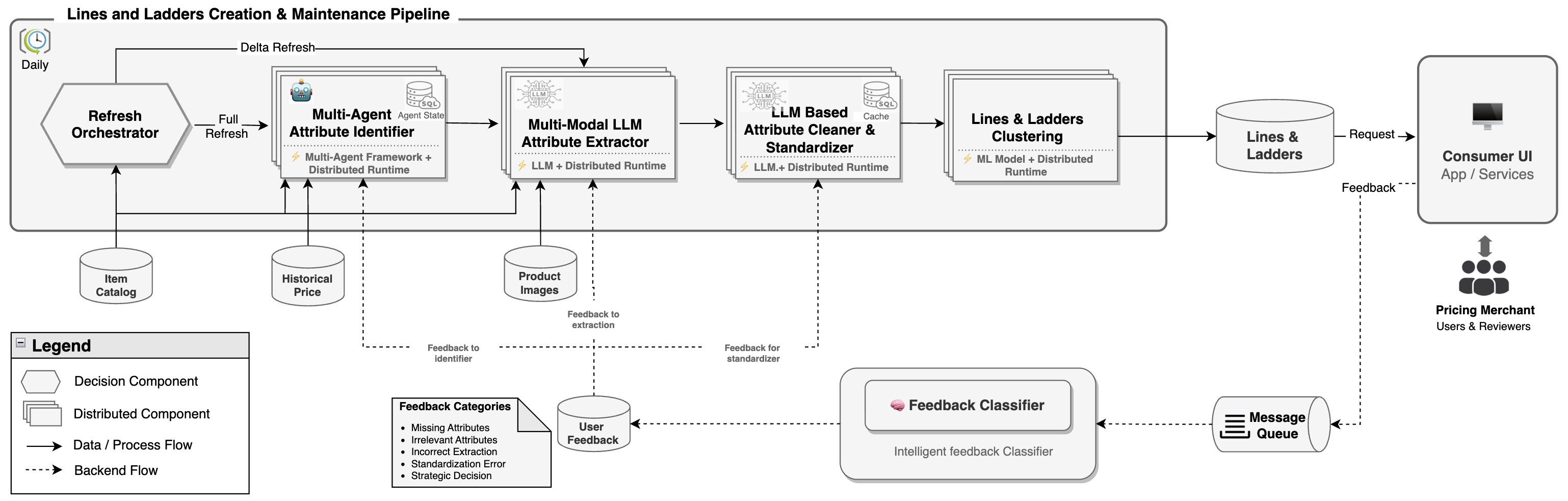}
\caption{Lines \& Ladder System Design}
\label{fig:System_Design}
\end{figure*}

Deployed on a distributed compute cluster for horizontal scalability and fault tolerance (Figure \ref{fig:System_Design}), the framework is managed by a refresh orchestrator supporting two execution modes: full refresh (recomputing attribute schema) and delta refresh (reusing persisted schema for steady-state updates).

\subsection{Distributed Pipeline Architecture}
Decomposing the pipeline by Department $\rightarrow$ \texttt{ProductType} enables independent failure domains. To balance reasoning and cost at scale, we employ a \textbf{multi-model routing strategy} using secure, enterprise-managed deployments of frontier commercial multi-modal LLMs:

\begin{itemize}
 \item \textbf{Multi-Agent Attribute Identifier:} Coordinates the three agents. Reasoning-heavy tasks (schema generation, conflict resolution) are routed to a proprietary, enterprise-managed frontier LLM (comparable in scale and capability to $>100B$ parameter instruction-tuned ``thinking'' models). A low-latency RDBMS implements persistent memory ($P_{context}$), using deterministic SQL queries to retrieve and inject historical merchant rules into prompts. To prevent context overflow, this injected history is strictly truncated to feedback collected since the last full refresh. Due to enterprise confidentiality, exact commercial model names cannot be disclosed. However, all agents utilize state-of-the-art frontier models with the temperature strictly set at 0 to maximize determinism.

 \item \textbf{Multi-Modal Attribute Extractor:} Executes extraction function $f_{LLM}$. To process millions of items, extraction is routed to a highly efficient, cost-optimized multi-modal model from the same family (comparable in scale to 7B-13B parameter Vision-Language Models), designed for high-throughput, low-latency execution.
 \item \textbf{Attribute Cleaner \& Standardizer:} Normalizes values via pipeline $\Psi$. A SQL-based semantic cache stores synonym mappings (e.g., ``Soy'' $\rightarrow$ ``Soy Wax'') to optimize latency and reduce redundant LLM calls during delta refreshes.
 \item \textbf{Lines \& Ladders Clustering:} Transforms standardized vectors into Lines ($L_k$) and Ladders ($M_j$) using hierarchical logic (Section 3.6.2).
\end{itemize}

\subsection{Asynchronous Feedback Loop}
Human validation utilizes an event-driven architecture. Merchant UI corrections emit events to a distributed message bus, where an Intelligent Feedback Classifier routes them by error category:
\begin{itemize}
\item Missing Attributes $\rightarrow$ Triggers agent refinement.
\item Extraction Errors $\rightarrow$ Updates Extractor prompts.
\item Standardization Errors $\rightarrow$ Updates SQL Cache.
\item Strategic Decisions $\rightarrow$ Updates memory $P_{context}$.
\end{itemize}
\subsection{Performance and Scalability}
We processed $\sim$1,700 \texttt{ProductTypes} ($\sim$1 million items) in 30 minutes using 10 worker nodes.
\begin{itemize}
\item \textbf{Throughput:} Distributed implementation enables linear worker node scaling within available LLM quotas.
\item \textbf{Cost Optimization:} LLM inference dominates spend ($\sim$4K tokens/item). Micro-batching and sampling during identification achieve significant cost reductions.
\item \textbf{Observability:} Structured logging enables partial reruns and \texttt{ProductType}-level regression analysis.
\end{itemize}
This architecture maintains strict catalog freshness SLAs within a sustainable cost envelope.

\section{Experimental Evaluation}
To validate the framework, we evaluate architectural ablation, structural accuracy, and post-launch business impact.
\subsection{Architecture Ablation}
To justify the multi-agent design, we conducted an ablation study across 14 categories. We specifically evaluate zero-shot LLM architectures rather than traditional supervised Named Entity Recognition (NER) models (e.g., fine-tuned BERT). Retail catalogs require dynamic schema generation (e.g., extracting ``wax\_type'' for candles but ``motor\_power'' for blenders); therefore, training and maintaining static models for thousands of evolving \texttt{ProductTypes} is operationally infeasible at enterprise scale. Furthermore, traditional dense embedding-based clustering approaches are inadequate for this task; they rely on semantic language similarity rather than the strict physical attribute equivalence and multi-modal business context required for pricing taxonomies. Consequently, our baseline comparison focuses on a Single-Agent LLM, a 2-Agent system (Similarity + Variance), and our proposed 3-Agent system.

\begin{table}[t]
\centering
\small
\renewcommand{\arraystretch}{0.8} 
\caption{Comparison across Agent Architectures}
\label{tab:ablation}
\resizebox{\columnwidth}{!}{%
\begin{tabular}{l ccc ccc}
\toprule
\multirow{2}{*}{\textbf{\texttt{ProductType}}} & \multicolumn{3}{c}{\textbf{Lines (F1)}} & \multicolumn{3}{c}{\textbf{Ladders (F1)}} \\
\cmidrule(lr){2-4} \cmidrule(lr){5-7} 
 & \textbf{1-Agt} & \textbf{2-Agt} & \textbf{3-Agt} & \textbf{1-Agt} & \textbf{2-Agt} & \textbf{3-Agt} \\
\midrule
Deodorants \& Antiperspirants & 0.63 & 0.90 & 1.00 & 0.63 & 0.90 & 1.00 \\
Prepared \& Packaged Soups & 0.60 & 0.92 & 0.96 & 0.60 & 0.79 & 0.92 \\
Juices & 0.66 & 0.69 & 0.90 & 0.69 & 0.69 & 0.92 \\
Ice Cream Bars, Cones & 0.50 & 0.70 & 0.88 & 0.73 & 0.70 & 0.95 \\
Over-the-Counter Medicine & 0.70 & 0.75 & 0.87 & 0.72 & 0.70 & 0.75 \\
Bath Soaps & 0.69 & 0.88 & 0.87 & 0.77 & 0.90 & 0.93 \\
Packaged Meals & 0.56 & 0.75 & 0.87 & 0.58 & 0.85 & 0.91 \\
Period Panties & 0.93 & 0.84 & 0.84 & 0.74 & 0.78 & 0.78 \\
Sandwiches, Filled Rolls & 0.73 & 0.80 & 0.82 & 0.82 & 0.84 & 0.95 \\
Toilet Paper & 0.54 & 0.52 & 0.79 & 0.67 & 0.31 & 0.76 \\
Laundry Detergents & 0.54 & 0.76 & 0.75 & 0.66 & 0.68 & 0.63 \\
Flushable Cleansing Cloths & 0.74 & 0.93 & 0.71 & 0.91 & 0.53 & 0.53 \\
Snack Crackers & 0.59 & 0.59 & 0.68 & 0.64 & 0.59 & 0.77 \\
Cat Food & 0.53 & 0.74 & 0.64 & 0.55 & 0.61 & 0.66 \\
\midrule
\textbf{Average} & \textbf{0.64} & \textbf{0.77} & \textbf{0.83} & \textbf{0.69} & \textbf{0.70} & \textbf{0.82} \\
\bottomrule
\end{tabular}%
}
\end{table}

As Table \ref{tab:ablation} shows, a single LLM tasked with simultaneous similarity and variance discovery suffers severe cognitive overload (0.64 Lines F1). While a 2-Agent system improves performance (0.77 F1), utilizing a Synthesis Agent (3-Agt) to resolve semantic overlaps was strictly necessary to achieve the $>80\%$ accuracy required for production.

\subsection{Quantitative Manual Evaluation}
To measure ``Assignment Accuracy'' (the percentage of items correctly grouped without requiring manual edits) in unstructured General Merchandise, domain experts evaluated 1000 randomly sampled items across 11 diverse \texttt{ProductTypes}. Because evaluating placement requires reviewing multi-modal data and historical pricing strategies, this represents tens of hours of specialized annotation. As shown in Table \ref{tab:GM_results}, the framework achieved an $80.0\%$ weighted average accuracy.

\begin{table}[b]
\centering
\caption{Manual Audit Results and Qualitative Feedback}
\label{tab:GM_results}
\begin{tabular}{lcc}
\hline
\textbf{\texttt{ProductType}} & \textbf{Sample Size} & \textbf{Accuracy} \\
\hline
Bean Bags & $100$ & $100\%$ \\
Face Mirrors & $38$ & $100\%$ \\
Air Conditioners & $100$ & $88\%$ \\
Surge Protectors & $100$ & $86\%$ \\
Humidifiers & $100$ & $84\%$ \\
Fertilizers & $100$ & $82\%$ \\
Sewing Machines & $100$ & $80\%$ \\
Ceiling Fans & $100$ & $72\%$ \\
Lamps & $100$ & $69\%$ \\
Office Boards & $62$ & $68\%$ \\
Duffel Bags & $100$ & $61\%$ \\
\hline
\textbf{Weighted Avg} & \textbf{1000} & \textbf{$80\%$} \\
\hline
\end{tabular}
\end{table}

Qualitative feedback reveals three distinct categories of error:
\begin{itemize}
 \item \textbf{Technical Granularity:} The model occasionally missed specific technical variants, such as ``Ultrasonic'' vs. ``Evaporative'' (Humidifiers), ``Down Rod'' vs. ``Hugger'' (Ceiling Fans), or ``Weed-Control'' vs. ``Feeding'' (Fertilizers).
 \item \textbf{Aesthetic \& Subjective Nuance:} Differentiation for Lamps or Duffel Bags often relies on intangible qualities (e.g., ``silhouette elegance'' or ``material quality'') requiring tacit domain knowledge, underscoring the need for HITL merchant intuition.
 \item \textbf{Data Completeness \& Complexity:} Errors in Sewing Machines correlated with sparse descriptions, while Office Boards struggled with complex bundle combinations (board + markers), confirming performance is bounded by catalog data quality.
\end{itemize}

Synthesizing these observations, in most failure cases, the grouping logic was directionally correct but missed a single differentiating attribute. This indicates the system does not hallucinate random groups, but rather is just one feature away from perfect alignment.

\subsection{Comparative Evaluation}
In Food \& Consumables, we utilized merchant-generated pricing structures across 14 \texttt{ProductTypes} as a directional benchmark, noting that merchants often construct lines based on ad-hoc strategies rather than strict attribute equivalence. We pre-processed reference data to remove statistical outliers.

\begin{itemize}
\item \textbf{Matching Logic \& Metrics:} Because our algorithm generates novel UUIDs, direct mapping to legacy IDs is infeasible. We utilized a maximum F1 score optimization approach (similar to bipartite matching) to align generated lines with merchant lines, calculating precision and recall.
\item \textbf{Results \& System Guardrails:} Table \ref{tab:food_results} shows an average \textbf{Line Precision of $98\%$} and \textbf{Line Recall of $75\%$} (Ladders: $92\%$ Precision, $81\%$ Recall). This high-precision/low-recall profile is an intentional guardrail. High-recall groupings risk applying price changes to unrelated items (e.g., incorrectly grouping 11 distinct ``Prego Sauces''). Our system conservatively groups only the 4 identical Alfredo sauces, safely protecting the customer experience.
\end{itemize}

\begin{table}[t] 
\centering
\small
\renewcommand{\arraystretch}{0.8} 
\caption{Comparative Performance against Merchants reference structures}
\label{tab:food_results}
\resizebox{\columnwidth}{!}{
\begin{tabular}{l ccc ccc}
\toprule
\multirow{2}{*}{\textbf{\texttt{ProductType}}} & \multicolumn{3}{c}{\textbf{Lines}} & \multicolumn{3}{c}{\textbf{Ladders}} \\
\cmidrule(lr){2-4} \cmidrule(lr){5-7} 
 & \textbf{F1} & \textbf{Prec.} & \textbf{Rec.} & \textbf{F1} & \textbf{Prec.} & \textbf{Rec.} \\
\midrule
Deodorants \& Antiperspirants & 1.00 & 1.00 & 1.00 & 1.00 & 1.00 & 1.00 \\
Prepared \& Packaged Soups & 0.96 & 1.00 & 0.93 & 0.92 & 0.93 & 0.93 \\
Juices & 0.90 & 1.00 & 0.85 & 0.92 & 1.00 & 0.88 \\
Ice Cream Bars, Cones & 0.88 & 1.00 & 0.83 & 0.95 & 1.00 & 0.92 \\
Over-the-Counter Medicines & 0.87 & 0.99 & 0.81 & 0.75 & 0.73 & 0.91 \\
Bath Soaps & 0.87 & 1.00 & 0.78 & 0.93 & 1.00 & 0.89 \\
Packaged Meals & 0.87 & 1.00 & 0.79 & 0.91 & 0.88 & 0.97 \\
Period Panties & 0.84 & 1.00 & 0.75 & 0.78 & 1.00 & 0.69 \\
Sandwiches, Filled Rolls & 0.82 & 0.95 & 0.76 & 0.95 & 1.00 & 0.91 \\
Toilet Paper & 0.79 & 0.97 & 0.71 & 0.76 & 0.82 & 0.78 \\
Laundry Detergents & 0.75 & 1.00 & 0.63 & 0.63 & 0.87 & 0.59 \\
Flushable Cleansing Cloths & 0.71 & 0.89 & 0.67 & 0.53 & 0.72 & 0.65 \\
Snack Crackers & 0.68 & 1.00 & 0.54 & 0.77 & 1.00 & 0.65 \\
Cat Food & 0.64 & 0.99 & 0.50 & 0.66 & 0.99 & 0.54 \\
\midrule
\textbf{Average} & \textbf{0.83} & \textbf{0.98} & \textbf{0.75} & \textbf{0.82} & \textbf{0.92} & \textbf{0.81} \\
\bottomrule
\end{tabular}%
}
\end{table}

As detailed in Table \ref{tab:food_results}, the framework consistently delivers near-perfect precision alongside a lower, category-dependent recall. This dynamic highlights a fundamental divergence between algorithmic strictness and human strategic intent:

\begin{itemize}
\item \textbf{High Precision:} The agent achieves near-perfect precision, confirming it groups physically identical items without hallucinating relationships, ensuring safe deployment.
\item \textbf{The Recall Gap:} Lower recall indicates the agent frequently over-splits lines compared to the merchant reference due to lacking strategic context. This explains the F1 variance: highly standardized categories (e.g., Deodorants, F1=1.00) align perfectly, whereas categories with subjective, marketing-driven descriptions (e.g., Cat Food, F1=0.64) rely on nuanced terms like ``Ocean Whitefish Pate'' vs. ``Flaked Tuna in Sauce''. Merchants frequently over-group these items based on promotional strategies or broad ``flavor families'' rather than strict attribute equivalence. Our algorithm's strict physical grouping naturally splits these broad merchant groups, resulting in lower recall.
\end{itemize}

This divergence highlights the necessity of the HITL workflow (Section 3.7). While the agent provides a safe, high-precision cold-start state, continuous merchant feedback enables the persistent memory to learn these strategic nuances, allowing future iterations to automatically align with the merchant's preferred granularity.

\subsection{Post-Launch Business Impact}
This fully launched production system processes tens of thousands of active items, generating structured Lines and Ladders for $>90\%$ of the previously un-managed catalog for the first time. Beyond creating this entirely new data foundation, end-to-end telemetry tracking merchant UI workflows over a 13-week period indicates an \textbf{$86.1\%$ reduction in time-on-task} for existing workflows, calculated via: $\text{New Hours} = \text{Old Hours} \times \text{Remaining Workload} \times \text{Remaining Time}$.
\begin{itemize}
 \item \textbf{Workload Reduction:} Automated extraction reduces manual catalog cleanup by $35-40\%$ (Remaining Workload $\approx 0.625$).
 \item \textbf{Velocity Increase:} Automated grouping and UI review is 4--5x faster than manual creation (Remaining Time $\approx 0.222$).
\end{itemize}
Applying this mid-case scenario ($0.625 \times 0.222 = 0.138$), the workflow requires only $13.9\%$ of original hours, enabling merchants to focus entirely on high-level pricing strategy.

\section{Limitations and Future Work}
\begin{itemize}
\item \textbf{Limitations: }Despite a temperature of 0, LLM non-determinism complicates strict reproducibility. The system is sensitive to sampling bias, where poor catalog quality can skew attribute discovery. Furthermore, compute costs currently restrict the framework to batch processing. In addition, reliance on specific LLM families necessitates re-validation upon model updates to mitigate hallucination risks. Finally, while we argue that traditional embedding-based clustering lacks the contextual reasoning required for this task, future work will include empirical benchmarking against these non-LLM baselines to rigorously quantify the multi-agent system's added value.
\item \textbf{Future Directions: }We plan to refine anchoring from \texttt{ProductType} to the brand level to capture niche features and utilize HITL feedback to stabilize non-deterministic outputs. We also aim to conduct longitudinal studies measuring the system's convergence to merchant strategy over multiple HITL feedback cycles, alongside statistical significance testing and sensitivity analysis on heuristic thresholds. To resolve data sparsity in niche categories ($\sim$5\%), we will implement cross-category transfer learning. Finally, to enable real-time inference, we will adopt a teacher-student architecture, distilling large models into Small Language Models (SLMs) for low-latency execution.
\end{itemize}

\section{Conclusion}
This paper presented ``Lines and Ladders'', a multi-agent framework that automates retail price governance by decomposing taxonomy construction into Similarity and Variance discovery. Synergizing LLM-based extraction with hierarchical grouping, the system structures heterogeneous data into coherent pricing tiers. 

Evaluations confirm $>80\%$ accuracy in General Merchandise and $>90\%$ precision in Food categories, with $75\%$ recall targeted for HITL optimization. Deployed in production, the system drastically reduces manual effort, enabling merchants to focus on high-level strategy. Beyond efficiency, this taxonomy establishes the structural foundation for autonomous pricing and anomaly detection. Ultimately, this work demonstrates that multi-agent architectures are highly viable for industrial-scale governance, offering a framework broadly applicable to other heterogeneous domains like industrial supply chains and online marketplaces.

\bibliographystyle{IEEEtran} 
\bibliography{references} 
\end{document}